\documentclass[sigconf]{acmart}

 \pdfoutput=1

\usepackage[utf8]{inputenc} %
\usepackage[T1]{fontenc}    %

\usepackage{color}
\usepackage{amssymb}
\usepackage{amsmath}
\usepackage[linesnumbered, ruled]{algorithm2e}
\usepackage[caption=false]{subfig}
\usepackage{graphicx}
 \usepackage{multirow}
 \usepackage{bbm}
 \usepackage{pgfplots}
 \usepackage{soul}
 \usepackage[font=small,skip=2pt]{caption}
 \usepgfplotslibrary{external} 
\newcommand\nb[1]{} %
\newcommand{\n}{^{(n)}} %
\newcommand\g[1][]{\:#1\vert\:} %
\newcommand{\nnnl}{\nonumber \\} %
\newcommand{\E}{\mathbb{E}} %

\usepackage{algorithm2e}
\graphicspath{{./}}

\title{Counterfactual Evaluation of Slate Recommendations with Sequential Reward Interactions}

\author{James McInerney$^{1\ast}$, Brian Brost$^{2\ast}$, Praveen Chandar$^{2\ast}$}
\author{Rishabh Mehrotra$^3$, Ben Carterette$^2$}\thanks{$^\ast$ Equal contribution authors.}

\email{jmcinerney@netflix.com; brianbrost,praveenr,rishabhm,benjaminc@spotify.com}
\affiliation{
\institution{$^1$ Netflix, Los Gatos, CA  $^2$ Spotify, New York, NY \\ $^3$ Spotify, London, UK}
}

\copyrightyear{2020} 
\acmYear{2020} 
\setcopyright{acmlicensed}\acmConference[KDD '20]{Proceedings of the 26th ACM SIGKDD Conference on Knowledge Discovery and Data Mining}{August 23--27, 2020}{Virtual Event, CA, USA}
\acmPrice{15.00}
\acmDOI{10.1145/3394486.3403229}
\acmISBN{978-1-4503-7998-4/20/08}

\begin{document}
\fancyhead{}
\begin{abstract}
Users of music streaming, video streaming, news recommendation, and e-commerce services often engage with content in a sequential manner. Providing and evaluating good sequences of recommendations is therefore a central problem for these services. Prior reweighting-based counterfactual evaluation methods either suffer from high variance or make strong independence assumptions about rewards. We propose a new counterfactual estimator that allows for sequential interactions in the rewards with lower variance in an asymptotically unbiased manner. Our method uses graphical assumptions about the causal relationships of the slate to reweight the rewards in the logging policy in a way that approximates the expected sum of rewards under the target policy. Extensive experiments in simulation and on a live recommender system show that our approach outperforms existing methods in terms of bias and data efficiency for the sequential track recommendations problem.
\end{abstract}

\maketitle

\section{Introduction}
\label{sec:Introduction}

Recommender systems enable users of online services to navigate vast libraries of content (e.g. news, music, videos). %
Evaluating such systems is a central challenge for these services, with A/B testing often regarded as the gold standard \cite{kohavi2013online, gomez2016netflix}.  
In A/B testing users are randomly assigned to different recommendation algorithms to isolate the treatment effects between them. 
Unfortunately, A/B tests are costly in several ways:
significant effort is required to implement new recommenders into production, the tests take weeks or months to run, and revenue is lost from sub-optimal user experiences if the existing recommender turns out to be better. %
Against this background, offline evaluation is crucial in rapidly validating new recommenders to ensure that only the strongest ideas are tested online.

Offline evaluation is challenging because the deployed recommender decides which items the user sees, introducing significant exposure bias in logged data \cite{Schnabel2016, gilotte2018offline, liang2016modeling}. 
Various methods have been proposed to mitigate bias using counterfactual evaluation. 
In this paper, we use terminology from the multi-armed bandit framework to discuss these methods: 
the recommender performs an \emph{action} by showing an item depending on the observed \emph{context} (e.g., user covariates, item covariates, time of day, day of the week) 
and then observes a \emph{reward} %
through the user response (e.g., a stream, a purchase, or length of consumption) \cite{li2010contextual}. 
The recommender follows a \emph{policy} distribution over actions by drawing items stochastically conditioned on the context.

The basic idea of counterfactual evaluation is to estimate how a new policy would have performed if it had been deployed instead of the deployed policy. 
This is challenging because it is easy to be fooled by spurious correlations between the reward and context in logged data. %
For example, logged data may indicate that pop music recommendations have a higher average reward but this effect may not hold up in intervention (i.e., through a new recommendation) if the production policy had been systematically recommending pop to younger users who may be more active on the service regardless of genre preferences.

The prevailing method for counterfactual evaluation is inverse propensity scoring (IPS). 
IPS approximates the average reward of a target policy offline by taking a weighted average of the rewards obtained under the production policy.  
The weights are a function of the ratio between target and production policies evaluated on the logged data. 
IPS estimators are usually unbiased (or asymptotically unbiased) but suffer high variance with large action spaces, a problem that is typical in recommendation. 
While simplifying assumptions can be made to reduce variance, they in turn introduce bias, particularly when there are interactions between rewards.
When recommendations are sequential, as in a playlist of tracks, these interactions sometimes dominate the overall reward. 
For example, a good song in an early position of a sequence can have a large effect on skip rates of songs at subsequent positions.

A major challenge in rapidly iterating on sequential recommendation systems is a reliable offline evaluation methodology. Sequential recommendations are becoming increasingly relevant due to richer interactions between users and recommenders~\cite{Guo2019,Ludewig2018,Tang2018,Yuan2019}. In such scenarios, the user's action (or reward) on a recommended item depends on the recent recommendations or user's previous actions. For example, a user purchasing a mobile phone is more likely to buy accessories in the near term. These effects are often not captured by traditional recommendation systems. %

In this paper, we introduce \emph{reward interaction inverse propensity scoring} (RIPS), a new approach to IPS estimation on sequences of sub-actions that avoids the modeling overhead of model-based estimation (i.e., the \emph{direct method}) and reduces the variance through structural assumptions about the reward dependencies in a slate. 
Our approach assumes a causal graph of conditional independencies of actions and rewards, enabling us to derive a counterfactual estimator that does not require distributional assumptions or parameter fitting.

In summary, our contributions are the following:
\begin{itemize}
	\item We formulate a new off-policy estimator of the total rewards of a slate 
	in counterfactual evaluation based on causal graph assumptions about 
	the interactions between the context, actions, and rewards in the slate. 
	The new estimator is a nested expectation 
	describing how reweighting factors accumulate down the slate.
	\item To approximate the nested expectation, 
	we propose a tractable algorithm~\footnote{The code for the proposed estimator is available at \url{https://github.com/spotify-research/RIPS_KDD2020}} that uses iterative normalization and lookback  
	to estimate the average reward of the target policy 
	from finite data collected using a logging policy. 
	\item Experiments show that the improved properties of RIPS 
	enable it to recover, with significantly greater accuracy than prior state-of-the-art, the ground truth value of the target policy both in simulation 
	and the outcome of A/B tests in a live sequential recommendation system. 
\end{itemize}

The rest of the paper is organized as follows. 
In Section~\ref{sec:RelatedWork}, we review related work in the area of counterfactual evaluation. 
In Section~\ref{sec:Counterfactual}, we present the RIPS algorithm. 
Sections~\ref{sec:ExperimentalEvaluation}~and~\ref{sec:realexperiments} describe the set of experiments in simulation and online (respectively) 
to compare RIPS against other baselines for counterfactual evaluation. 
Finally, we conclude and discuss future work in Section~\ref{sec:Conclusion}.

\section{Related Work}
\label{sec:RelatedWork}

In this section, we discuss prior work on evaluating recommender system using implicit and explicit feedback both online and offline.

\subsection{Online evaluation}
The most trusted method for measuring the quality of a recommendation system, conventional or sequential, is using online controlled experiments.  
In this setup, comparing two or more recommendation algorithms involves redirecting the user traffic to different variants~\cite{kohavi2013online} and measuring the quality of recommendation algorithm, often based on a predefined user satisfaction metric.  
In this work, we rely on a simple A/B test setup to estimate the true rewards of different policies in our experiments.

\subsection{Offline evaluation}\label{offlineeval}
While online evaluation is a reliable method of evaluation, it takes several weeks to collect sufficient amount of data to reliably compare systems. 
Further, there is the risk of user attrition since we could expose potentially bad recommendations to real users. 
On the other hand, offline evaluation is cheap and enables us to test ideas rapidly without exposing experimental systems to live traffic. 

\subsubsection{Traditional Methods}
For several decades, recommender and search systems evaluations have heavily relied on offline test collections for rapid experimentation~\cite{Sanderson2010}; however, they are expensive and require a considerable amount of time and effort to curate. Furthermore, such collections cannot easily be adapted to account for personal or time-sensitive relevance.
Alternatively, the use of explicit feedback such as user ratings has been explored, but prior research has pointed out that explicit feedback information collected is often missing information, not at random (MNAR)~\cite{Schnabel2016, Steck2013, Marlin2009} leading to bias in evaluation.

Implicit signals such as clicks are a rich source of information for offline evaluation, since they are collected in a natural setting, typically reflecting the user's preferences and available at a low cost. 
However, recent work has pointed out that the implicit data collected is subject to different sources of biases~\cite{Schnabel2016}. 
Specifically, the feedback collected leads to sampling bias since the data collection is mediated by the recommendation system themselves; our focus is on addressing these biases in this work.

\vspace{-1em}
\subsubsection{Counterfactual Methods}
Counterfactual analysis techniques are increasingly used for training and evaluating machine learned recommendation models
from user log data collected online.
One specific technique is {\em inverse propensity scoring} (IPS) \cite{nedelec2017comparative}, which has a long history in the study of experimental design and clinical trials. IPS is a way to estimate the average reward of a target policy from data collected according to a logging policy while correcting for the mismatch between target and logging policy.

IPS estimators can be proven to be unbiased under weak assumptions. But even when those assumptions are true, IPS estimators suffer high variance with large action spaces~\cite{gilotte2018offline}. Large action spaces occur when there are many items in the catalog and when recommendation happens in slates. Slate recommendation is an increasingly important and common interface design that presents a set of items to a user at once (e.g., search results pages, a shelf of videos, a playlist of tracks) 
and the user may engage with any or none of the items or {\em sub-actions} in the slate \cite{swaminathan2017off}.  
The sub-actions together affect the observed reward, 
making it difficult to disentangle the role of any particular sub-action. 
There are a combinatorially large number of potential actions 
resulting in a very small overlap between the production policy and target policy. 
Consequently, the weights in IPS are mostly zero, ignoring most of the data 
and giving a high variance estimate of the target policy reward. 

\textbf{\emph{Variance Reduction Methods:}} Capping IPS weights and normalizing them against their sum has been shown to reduce variance, but increase bias \cite{gilotte2018offline}.
In other cases, simplifying assumptions can be deployed to reduce the variance 
of IPS estimators in slate recommendation. 
Li et al. assume that the reward for each sub-action is independent of other sub-actions in the slate \cite{li2018offline}; 
this is a typical assumption to greatly reduce the size of the action space 
that is used in practical applications \cite{chen2019top}. 
To simplify slate reward estimation, 
Swaminathan et al. take an additive approach to the slate reward 
and assume that the sub-action rewards are unobserved and independent~\cite{swaminathan2017off}. 
In contrast, our method assumes that the sub-action rewards are observed and we do not assume independence.
Looking at other problem settings, different schemes for lookback capping have been proposed to reduce variance for off-policy estimation in reinforcement learning \cite{thomas2016data} and to reduce bias in the non-stationary contextual bandit setting \cite{jagerman2019people}.

\textbf{\emph{Model-Based or Direct methods:}} 
In general, with the exception of naïve IPS, which suffers from excessive variance, 
the bias introduced by existing IPS methods 
is large when the rewards for sub-actions in a slate are not independent.
The only existing general approach to taking advantage 
of sequential structure in the rewards of sub-actions 
in a slate is the direct method \cite{dudik2011doubly}. 
In the direct method, a model of rewards is posited 
then trained on logged data. 
The direct method requires a 
significant amount of modeling overhead 
to specify the distribution of rewards, 
fitting their parameters from data, 
criticizing the model, and finding the right hyperparameters. 
This effort does not translate across datasets/applications and is vulnerable to overfitting. 
Doubly robust methods are well known to reduce the variance of IPS estimators \cite{dudik2011doubly}. 
but carry all the overhead of the direct method, and, 
due to the combinatorial explosion in the number of possible slates, 
even a good control variate 
will not alleviate the aforementioned issues with~IPS. Related methods that combine multiple sub-estimators such as \cite{thomas2016data}, and generalizations thereof \cite{su2019cab} can improve the bias-variance tradeoff, but do not eliminate the aforementioned modeling overhead. An interesting future exploration could be to characterize how our estimator can be incorporated into these combined estimator families.

\section{Reward Interaction Inverse Propensity Scoring (RIPS)}
\label{sec:Counterfactual}

   \begin{figure*}[t!]
   \centering
     \subfloat[Standard IPS\label{fig:IPS}]{%
       \includegraphics[width=0.25\columnwidth]{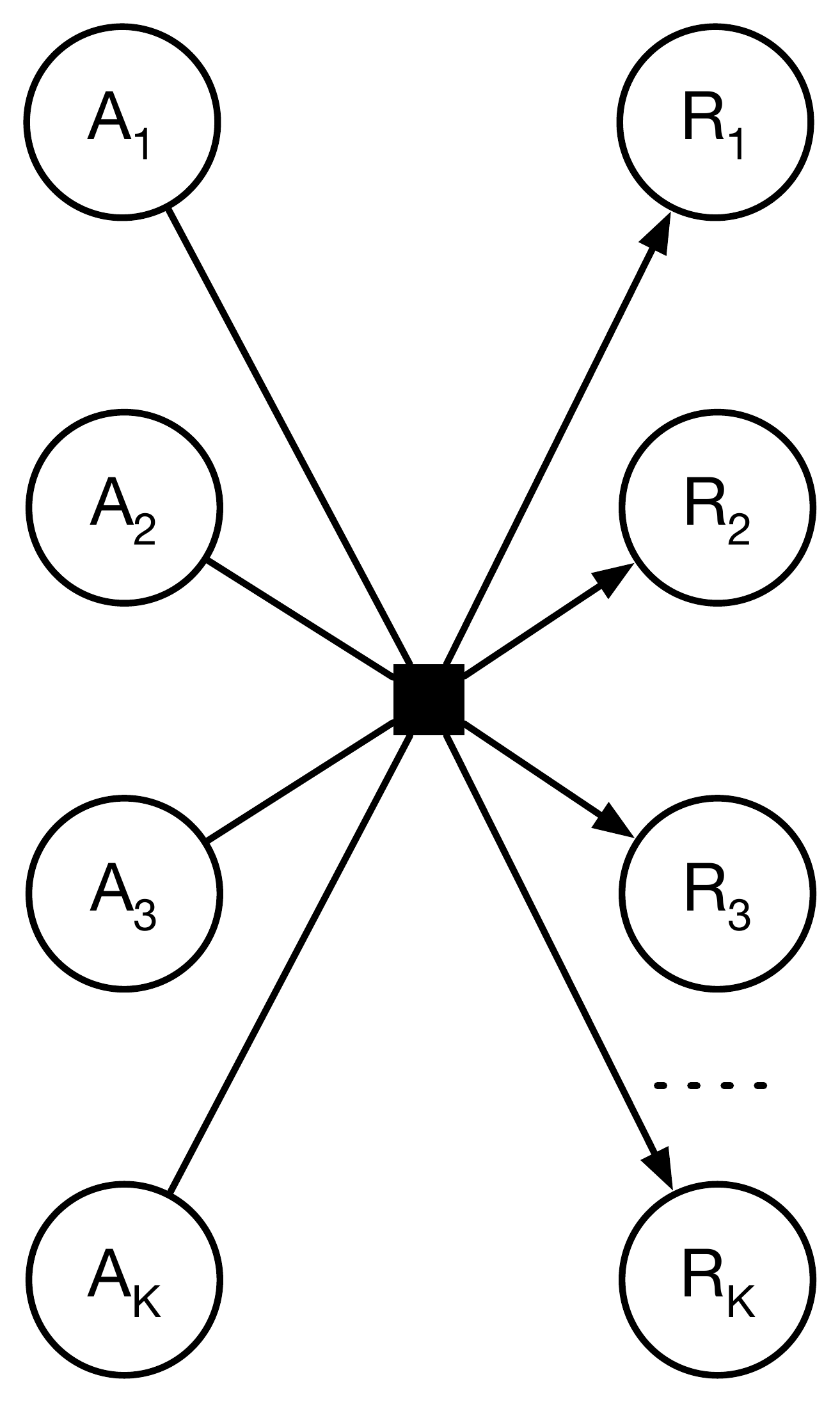}
       \label{fig:ips_gm}
     }
     \hspace{50pt}
     \subfloat[Independent IPS and PI]{%
       \includegraphics[width=0.25\columnwidth]{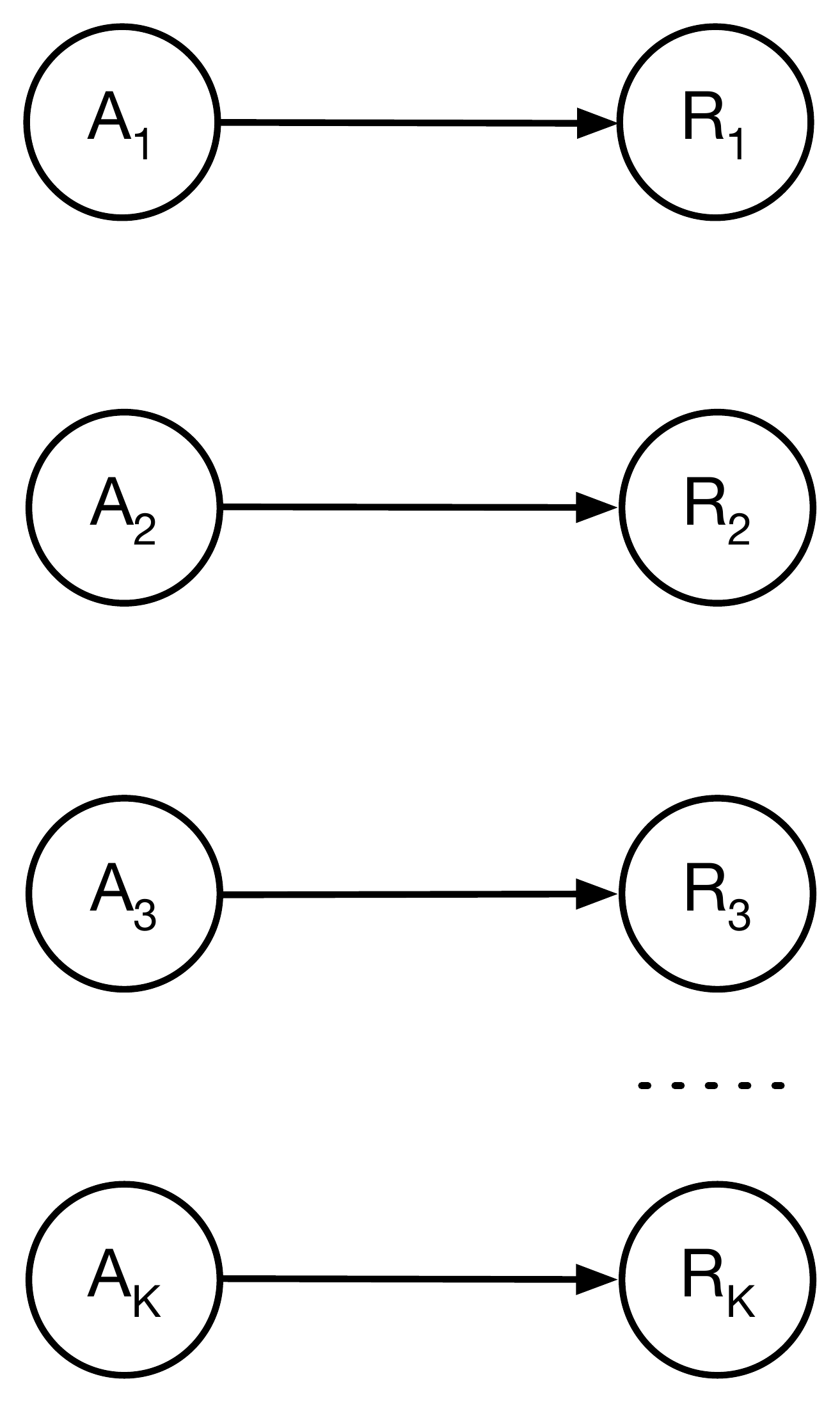}
       \label{fig:iips_gm}
     }
     \hspace{50pt}
     \subfloat[Reward interaction IPS (RIPS)\label{fig:RIPS}]{%
       \includegraphics[width=0.25\columnwidth]{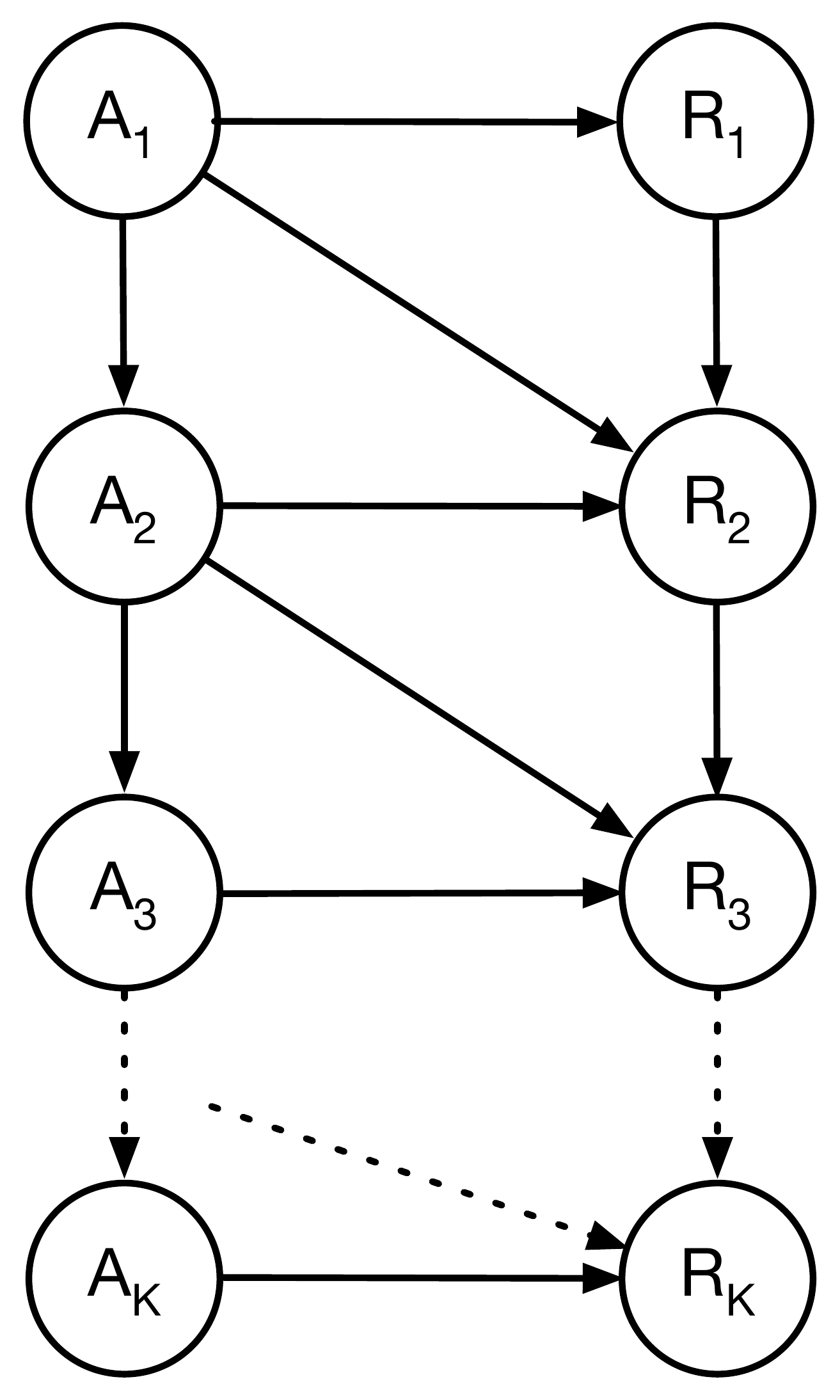}
       \label{fig:rips_gm}
     }
     \caption{The assumptions of each IPS-based estimator described in this work can be visualized as a causal graphical model.}
     \label{fig:graphical}
   \end{figure*}

In this section, we describe the problem of slate recommendation 
and discuss the shortcomings of inverse propensity scoring and its variants. 
Using causal assumptions about the actions and rewards in the slate, 
we present our algorithm for counterfactual~evaluation.

\subsection{Problem Formulation}
\label{sec:problem_formulation}

Slate recommendation posits a sequence of sub-actions $A_{1:K}$ 
chosen by a policy $\pi$ 
in a context $X$ 
that results in a total reward $R$ for the slate. 
In the scope of this work, 
we consider the setting where 
there is an observed reward $R_k$ 
for each ordered sub-action $A_k$ in the slate 
and the total reward is the sum of the sub-action rewards, 
i.e., $R = \sum_{k=1}^K R_k$.\footnote{Our derivations are trivial to extend to the more general case that the total reward for a slate is an affine function of the sub-action rewards. We focus on the sum of sub-action rewards for simplicity of exposition.}
We do not assume 
that the rewards are independent. 

Suppose $\pi$ is a recommender 
that takes contextual features about a user 
(e.g. device, time of day, long-term user features) 
and %
returns an ordered set of items. 
The overall reward for the slate is 
the total number of engagements (e.g. purchases, clicks, completions) 
and these engagements, 
though correlated, 
are associated with individual actions. 
An example of reward correlation 
occurs in playlist listening: 
if a user hears a song they dislike in a playlist, 
they may skip it 
and may then be more likely to skip the next song. 
In other settings, e.g., homepage recommendation, 
there is a negative correlation between 
the user response to items high in the list 
and those lower in the list 
due to a limited attention budget. 

When run in production, 
$\pi$ produces slates and logs impressions. 
We use superscript $n$ ($1 \le n \le N$) 
to refer to the realization of the random variable 
on impression $n$, 
e.g., $R_k\n$ is the reward for the $k^\mathrm{th}$ action $A_k\n$ of the $n^\mathrm{th}$ impressed slate $A^{(n)}$. 

After running $\pi$ in production, 
the feedback logs will consist 
of all the sampled contexts, actions, and rewards 
$\left\{ \left(X\n, A_{1:K}\n, R_{1:K}\n \right)_{n=1}^N \right\}$. 
In general, the size of the slate $K$ may be different 
across impressions 
and there may be a context for every position in the slate. 
For simplicity of notation, we use constant $K$ and $X^{(n)}$ across impressions, though this assumption is not required by our method. 

The goal of counterfactual evaluation 
is to estimate the average reward for a target policy $h$. 
All methods in counterfactual evaluation 
estimate $V(h) = \E[ \sum_{k=1}^K R_k]$ 
using the logged data. 
Our starting point 
in addressing this problem 
is a popular class of evaluators 
based on importance sample reweighting. 

\subsection{Inverse Propensity Scoring (IPS)}

Inverse propensity scoring (IPS) methods 
use the importance sample reweighting method 
to evaluate a target policy $h$ 
by reweighting the rewards received from collection policy $\pi$, 
\begin{flalign}
	\hat{V}_\mathrm{IPS}(h) = \frac{1}{N} \sum_{n=1}^N \frac{\prod_{k=1}^K h(A_k\n \g X\n)}{\prod_{k=1}^K \pi(A_k\n \g X\n)} \sum_{k=1}^K R_k\n \label{eq:ips} 
\end{flalign}
where $X\n \sim p(X)$, $A\n \sim \pi(A \g X\n)$, $R\n \sim p(R \g X\n, A\n)$, and $p(X)$ and $p(R \g X\n, A\n)$ 
are the unknown distributions for the context and rewards.\footnote{N.B. all IPS-based methods discussed in this paper admit policies that condition on previous actions. This is possible because IPS interacts with known functions $\pi$ and $h$ via a score for each sub-action that may depend on the context and any previous sub-actions in the slate.}

In this work, 
we examine 
IPS-based methods 
and their properties using the causal graph representation 
of Pearl's $do$-calculus \cite{pearl2009causality}. 
A causal graph expresses conditional independencies 
between random variables 
in the same way a probabilistic graphical model does. 
In addition, 
the $do(X=x)$ operator 
on a random variable $X$ 
represents an intervention to make $X=x$ 
that removes all incoming edges in the graph and forces downstream variables to condition on $X=x$. 
This approach allows us to formulate 
novel IPS methods 
to take advantage of sequential structure in the slate rewards.

IPS usually requires the stable unit treatment value assumption (SUTVA) in order to derive an estimator that reweights individual actions (i.e., units) \cite{imbens2015causal}. 
The individual actions in slate recommendation violate SUTVA 
because the action selected in one position affects the outcome of actions in other positions. 
The typical strategy to address SUTVA violations is to change how actions are defined. 
Specifically, one can redefine actions in slate recommendation as the ordered sequence of sub-actions. 
The issue with doing so is that the action space increases combinatorially 
resulting in high variance with IPS. 
For this reason, standard IPS is limited in its applicability 
to slate estimation.

 \subsubsection{Independent IPS} 
In order to widen the applicability of IPS-methods 
to larger slates with different $h$ and $\pi$, 
existing work uses an item-position click model 
that applies IPS under the assumption 
that the probability of a click 
depends only on the item and its position~\cite{li2018offline}, 
\begin{flalign}
	V_\mathrm{IIPS}(h) &= \E_X[ \E_{A \sim h(\cdot \g X), R \sim p(\cdot \g X, A)}[ \sum_{k=1}^K R_k] ] \nnnl
	&= \{ \mathrm{assume} \; R_k \; \mathrm{depends\;only\;on}\; X, \; k, \;\mathrm{and\;} A_k \} \nnnl 
	& \;\;\;\;\;\E_{X, A \sim h(\cdot \g X)}[ \sum_{k=1}^K \E_{R_k \sim p(\cdot \g X, A_k)}[ R_k] ] \nnnl
	&= \E_{X, A \sim \pi(\cdot \g X)}[\sum_{k=1}^K \E_{R_k \sim p(\cdot \g X, A_k)}[  \frac{h(A_k \g X)}{\pi(A_k \g X)} R_k] ] \nnnl 
	&\approx \frac{1}{N} \sum_{n=1}^N \sum_{k=1}^K w_k^{(n)} R_k^{(n)}, 
	\mathrm{\;\;where\;} w_k^{(n)} = \frac{h(A_k^{(n)} \g X^{(n)})}{\pi(A_k^{(n)} \g X^{(n)})} \label{eq:IIPS}
\end{flalign}
We refer to this estimator as \emph{independent IPS} (IIPS) 
because the rewards are assumed independent 
of any other actions or rewards in the slate. 
IIPS is a more general version of the first-order approximation of IPS that treats positions identically \cite{achiam2017constrained, chen2019top}. 
We focus on IIPS here because we found empirically that it has sufficiently low variance. 
We show the independence assumptions 
for IPS and IIPS 
graphically in Figures~\ref{fig:ips_gm}~and~\ref{fig:iips_gm}. %
Nodes indicate random variables and 
arrows represent direct dependencies between random variables. 

While IIPS has proven effective at 
counterfactual slate reward estimation, 
it assumes that each reward in the slate 
is isolated from every other action and reward. 
This rules out any sequential interactions 
between actions and rewards in the data. 
If this does not hold, 
the estimate will be biased 
no matter how much data is collected. 
Furthermore, the estimator becomes increasingly confident 
in the biased estimate with more data.

\subsubsection{Pseudoinverse Estimator}\label{ss:pi}
The pseudoinverse estimator (PI) for off-policy evaluation~\cite{swaminathan2017off} 
assumes that the observed slate-level reward is 
a known linear function of the sub-action rewards, 
and that the sub-action rewards are not observed. 
These, 
along with the absolute continuity assumption
(i.e. that $h(A \g X) > 0 \implies \pi(A \g X) > 0, \;\forall A, X$),  
allow the off-policy evaluation problem 
to be restated as a linear regression problem, 
where the sub-actions and their positions are 
encoded in a feature vector $\mathbbm{1}_A$ 
of ones and zeros (depending on slate $A$) 
and a set of unknown weights that correspond to the sub-action rewards, 
\begin{flalign}
	\hat{V}_\mathrm{PI}(h) &= \frac{1}{N} \sum_{n=1}^{N} \E_h[\mathbbm{1}_{A^{(n)}} \g X^{(n)}] ^\top\E_\pi[\mathbbm{1}_{A^{(n)}} \mathbbm{1}_{A^{(n)}}^\top \g X^{(n)}]^{\dagger}\mathbbm{1}_{A^{(n)}}  \sum_{k=1}^K R_k^{(n)}, \nonumber %
\end{flalign}
where superscript $\dagger$ indicates the pseudoinverse of a matrix. 

PI is not well suited to situations where the sub-action rewards are observable, 
i.e., when the credit for the success of a slate 
is attributable to a particular sub-action. 
Further, the linearity assumption means that 
the rewards for sub-actions are assumed independent. 
To loosen the independence assumptions of both IIPS and PI 
to produce a reliable estimate, 
we next introduce an
extension to IPS
that deals with sequences of actions and rewards 
in a slate.

\subsection{Reward interaction IPS (RIPS)}

Reward interaction IPS (RIPS) 
assumes a Markov structure 
to the rewards 
as shown in Fig.~\ref{fig:rips_gm}. 
The reward at position $k$ 
is directly conditional on the action performed at $k$, 
and the previous action and reward at position $k-1$. 
The dependence of rewards on each other 
makes it very challenging to estimate the expected off-policy reward 
because the usual trick of replacing the expectation 
with an empirical average is not available in this case. 
Furthermore, the downstream effects of an intervention at position $k$ 
could extend beyond $k+1$ through cascading effects. To motivate 
the assumption underlying RIPS, we note that for the real-world
music streaming data introduced in Section~\ref{sec:realexperiments}, the average
skip rate is 0.41, but the skip rate given a prior skip is 0.78, and the skip rate given the previous track was not skipped is 0.21. 
This kind of sequential dependence is not taken into account by IIPS and PI. 

Note that, 
although we specified conditional independence assumptions, 
we have not posited a full reward model 
that describes the effect of actions on rewards. 
(If we were to do this, we would be using the direct method.)\footnote{
For example, specifying that the reward $R_k$ 
is a parametric function of $A_k$ and $A_{k-1}$ 
and using this predictor to directly calculate the value of another policy.} 
The advantage of IPS methods 
is that they avoid 
the additional burden of fitting parameters 
and dealing with the bias that action-reward model 
assumptions introduce.

In the first instance, 
we can apply the conditional independence assumptions 
to refine the IPS estimate into a nested expectation,
\begin{flalign}
	V_\mathrm{Nested}(h) &= \E_X[ \E_{A \sim h, R}[ \sum_{k=1}^K R_k] ] \nnnl
	&= \{ \mathrm{assume} \; R_k \; \mathrm{depends\;only\;on}\; X,\; k,\; A_k,\,\mathrm{and}\;R_{k-1}\} \nonumber \\ 
				& \quad \E_{X} [ \E_{A_1 \sim h, R_1} [ R_1 + \E_{A_2 \sim h, R_2}[ R_2 + \dots \nonumber\\
				& \quad + \E_{A_K \sim h, R_K}[ R_K \g R_{K-1} ] \g R_{K-2}] \dots \g R_1] ]], \label{eq:nested_exp}
\end{flalign}
where we suppressed some notation for readability.
Applying importance sample reweighting results in nested reweightings 
with the same structure, 
under the usual assumption of absolute continuity, 
\begin{flalign}
	V_\mathrm{Nested}(h) &= \E_{X} [ \E_{A_1 \sim \pi, R_1} [ \frac{h(A_1 \g X)}{\pi(A_1 \g X)} ( R_1
	 + \E_{A_2 \sim \pi, R_2}[ \frac{h(A_2 \g X, A_1)}{\pi(A_2 \g X, A_1)} \nnnl ( R_2 + 
	 & \quad \dots + \E_{A_K \sim \pi, R_K}[ \frac{h(A_K \g X, A_{K-1})}{\pi(A_K \g X, A_{K-1})} R_K \g R_{K-1} ]) \g R_{K-2}]) 
	\nnnl & \quad \dots \g R_1]) ]]. \label{eq:RIPS}
\end{flalign}
Eq.~\ref{eq:RIPS} indicates that extreme propensity scores are still a problem, 
as the $j^\mathrm{th}$ term will have a weight of $\prod_{k=1}^j \frac{h(A_k \g X, A_{k-1})}{\pi(A_k \g X, A_{k-1})}$. To address this, 
we make use of normalization and \emph{lookback capping}.

In more detail, 
the expected reweighting factor for any subsequence of $k$ actions in the slate is 1 
due to, 
\begin{flalign}
	\E \left[ \frac{h(A_{1:k})}{\pi(A_{1:k})} \right] = \int \frac{h(A_{1:k})}{\pi(A_{1:k})} \pi(A_{1:k}) \mathrm{d}A_{1:k} = \int h(A_{1:k}) \mathrm{d}A_{1:k} = 1. \label{eq:derive_1}
\end{flalign}
This leads to an iterative normalization procedure for 
approximating the nested expectation 
in Eq.~\ref{eq:nested_exp} 
based on accumulating a message $\gamma_k$ 
through the reward chain 
such that 
$\gamma_k$ is the reweighting factor for the reward at position~$k$:

\begin{flalign}
	\hat{V}_\mathrm{RIPS}(h) &:= \frac{1}{N} \sum_{n=1}^N \sum_{k=1}^K \gamma_k^{(n)} R_k^{(n)}, \label{eq:recursive_normalizer}\\
	&\mathrm{where\;} \gamma_k^{(n)} := \frac{N \gamma_{k-1}^{(n)} w_k^{(n)}}{\sum_{n'=1}^N \gamma_{k-1}^{(n')} w_k^{(n')}} 
	\mathrm{\;and\;} \gamma_0^{(n)} := 1. \label{eq:def_Y}
\end{flalign}

The reweighting factors for actions toward the end of the slate 
are still a product of a large number of weights (that are often close to zero)
and normalization addresses but does not solve the problem. 
To further mitigate extreme weights, we propose a form of \emph{lookback capping} 
to limit the effect of combining more weights than the data support. 
Lookback capping uses the effective sample size (ESS) \cite{kish1965survey} 
to estimate the amount of data that are available to provide an estimate at position $k$ 
based on the overlap between the logging and target policy, 
\begin{flalign}
	ESS(\gamma_k^{(1:N)}) = \frac{N^2}{\sum_{n=1}^N (\gamma_k^{(n)})^2} \label{eq:ESS}
\end{flalign}
Intuitively, as we increase the lookback at position $k$, 
the weights will become increasingly extreme, 
causing the effective sample size to fall below some threshold $t$ that represents a 
proportion of the size of the dataset.
We choose $t$ to balance the number of interactions considered in the estimator 
against the high variance introduced by extreme weights. 
The lookback should be as high as possible while also ensuring ESS $> N t$.

The reason why it is not necessary to look forward when computing reweighting factors 
is because the outcome at position $k$ is only affected by 
$do$-calculus interventions from position $1$ to position $k$. 
This means that, in contrast to sequential inference in non-causal graphs, 
we only need to do filtering and no smoothing. 

The full RIPS algorithm is presented in Algorithm~\ref{alg:RIPS} in Appendix~\ref{appendix:rips}. 
It has time complexity $\mathcal{O}(NK^2)$, where $K$ is the slate size (usually a small constant). 
A test on line~12 of the algorithm 
is added to ensure that the ESS is always decreasing as $b_k$ increases. 
In practice, the ESS increases after a large enough lookback 
because the more weights that are included, the greater the chance that 
every sample contains an extreme weight, 
causing normalization to weight the samples equally. 
This raises the ESS despite the fact that every sample is now equally bad.

When the conditional independence assumptions 
shown in Figure~\ref{fig:RIPS} hold, 
RIPS has lower variance than standard IPS (and its normalized variant NIS \cite{gilotte2018offline}) 
and is both consistent and asymptotically unbiased. 
More details about the properties of RIPS are provided in Appendix~\ref{appendix:rips}.

\section{Simulation Experiments}
\label{sec:ExperimentalEvaluation}

\begin{figure*}[t]
\begin{center}
\centering 
\includegraphics[scale=0.4]{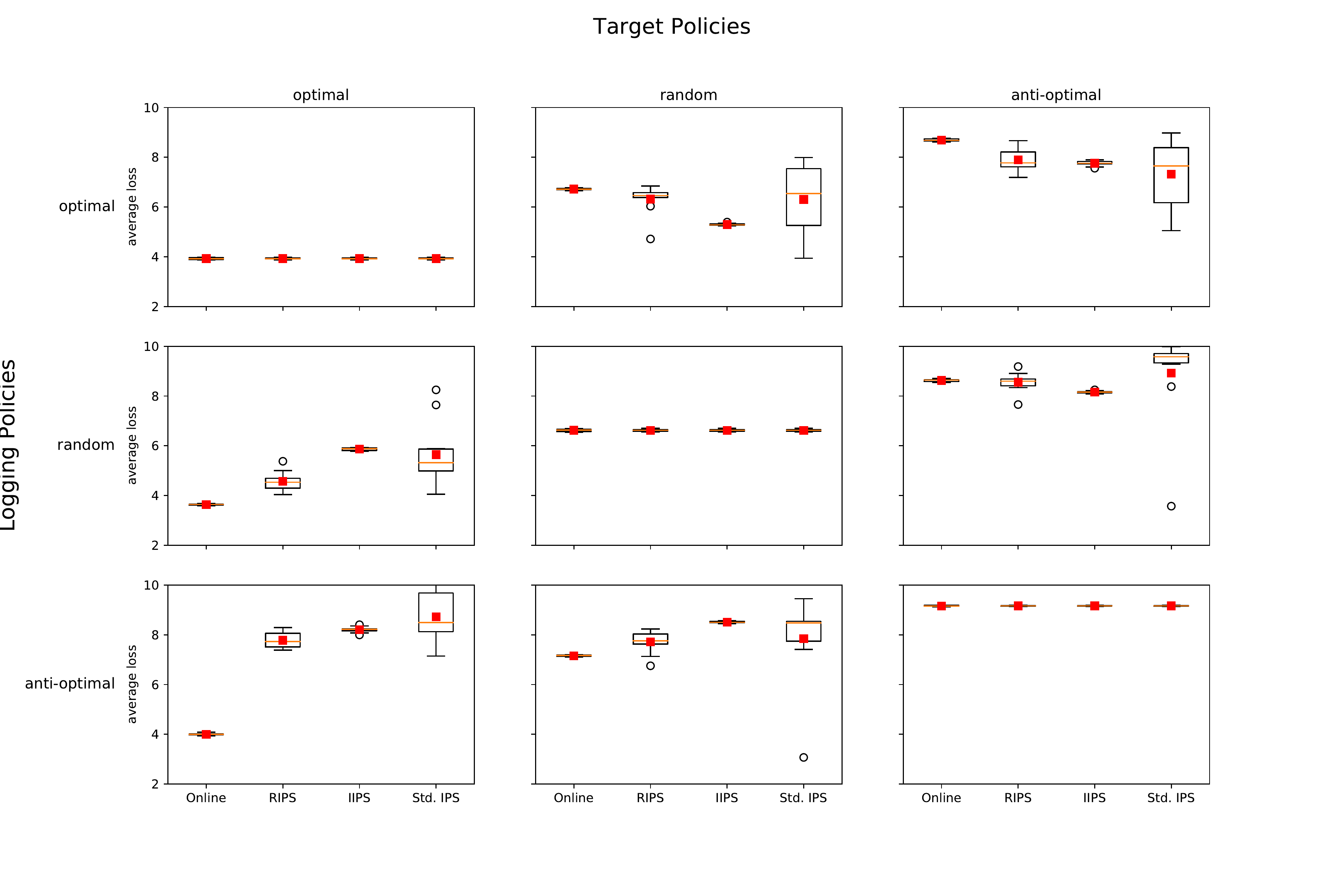}
\vspace{-10pt}
\caption{\emph{Top}: Results of the simulation experiment showing the average reward estimates for three different target policies computed using various off-policy estimators when using optimal policy as the logging policy. \emph{Middle:} same as the top but with uniform random as the logging policy. \emph{Bottom:} same as the top but with anti-optimal policy as the logging policy.}
\label{fig:plot1}
\end{center}
\end{figure*}

We adopt a dual approach to empirically evaluate the performance of our proposed reward interaction inverse propensity scoring estimator. 
First, we use simulation experiments to illustrate how existing off-policy estimators are biased, or suffer from extreme variance, when the user's rewards are dependent on previous rewards. We also demonstrate the ability of our proposed estimator to deal with such reward interactions. 
In Section~\ref{sec:realexperiments}, we show the practical applicability of our problem setting and approach by testing how the estimators evaluate different recommendation policies on user data collected during an A/B test.

\noindent \subsection{Simulation Setup}
\label{sec:syntheticexperiments}
In this section, we describe the simulation setup we use to illustrate how simplifying assumptions made by existing off-policy estimators lead to biased estimates. Specifically, we show that in the presence of interactions between rewards, an off-policy estimator must explicitly account for the interactions to provide unbiased estimates.  We designed experiments around simulations of contexts, slates, and user interactions on a given slate; this allowed us to analyze the counterfactual methods under a flexible and controlled environment.  We briefly describe the components of our simulation setup below and report the results of various state-of-the-art estimators.

\subsubsection{Context and reward simulation:} \label{sec:exp:reward_simulation}
We generate simulated contexts, each consisting of a set of candidate sub-actions.  In this study, we use ten sub-actions for a given context. This setting reflects a typical sequential recommendation scenario. For example, consider the task of recommending tracks within an algorithmic playlist in online music streaming services where the context includes details of about user and the sub-actions are the tracks to be recommended. Next, we randomly assign the true reward (i.e., probability of streaming a track in the case of track recommendation) for the sub-actions in the candidate pool for each context. Note that the true rewards are not known to the off-policy estimators in the simulation. 

\subsubsection{User Simulation:}
We simulate a user examining a slate sequentially starting from the top then examining each sub-action at a time. The rewards on the sub-actions are assumed to be binary, i.e., the user provides a positive or negative reward for each sub-action. Further, the simulation assumes reward interactions, i.e.,  the reward on a sub-action is conditional on previous rewards and sub-actions on a given slate. In this work, we used a reward model that has a cascading effect. When a user provides a negative reward to a sub-action, it has a cascading effect resulting in a negative reward for the next sub-action in the slate. This user model is similar to the one used by Chandar and Carterette~\cite{Chandar2018}. We choose a more extreme version of cascade behavior to illustrate the bias introduced by such a user behavior in existing estimators. 

Note that our assumptions about the user model will affect the estimates provided by the counterfactual estimators. Therefore, we rely on real-world experiments (see Section~\ref{sec:realexperiments}) to validate these assumptions about the user model and in turn, validate the comparison of estimators.

\subsubsection{Policy Simulation:}
We simulate three simple policies in our experiments: \emph{optimal}, \emph{anti-optimal}, and uniform random policy. Since the true reward for all the sub-actions for a context is known ahead of time, we simulate the \emph{optimal} and \emph{anti-optimal} policy by ordering the sub-actions based on true rewards as generated from Section~\ref{sec:exp:reward_simulation}. In other words, we sort the sub-actions (i.e., tracks) by the true reward (i..e, probability of stream) to get the optimal policy and reverse sort to get the anti-optimal policy. We also experimented with different policies with varying degrees of true effectiveness and observed similar results.

\subsection{Off-Policy Estimators}
\label{sec:exp:offpolicy}
In our experiments, we compare our proposed off-policy estimator to various state-of-the-art method in the literature that are relevant to our problem. We briefly describe the estimators compared below:
\begin{itemize}

\item {\bf Standard IPS}: takes the whole slate as an action and estimates the average reward. In our experiments, we use normalized importance sampling in which the weights are rescaled to sum to 1 in order to reduce variance \cite{swaminathan2015self}. 

\item {\bf Independent IPS (IIPS)}: Proposed by Li et al.~\cite{li2018offline}, the estimator treats the reward at each position independently. The average reward of the target policy is estimated using Eq.~\ref{eq:IIPS}.

\item {\bf Reward interaction IPS (RIPS)}: estimates the nested expectation by accumulating weights from the top of the slate to the bottom and estimates the average reward using Eq.~\ref{eq:RIPS}.
\end{itemize}

\subsection{How does reward interaction in user behavior affect off-policy estimation?}

We evaluated the effectiveness of the three simulated policies---optimal, anti-optimal, and uniform random---using different off-policy estimators described in Section~\ref{sec:exp:offpolicy}.  
Figure~\ref{fig:plot1} shows the estimated average reward for the three target policies while using the logs from the different logging policies.
The three rows in the figure represent the same experiment repeated with different policies as the logging policy, the diagonal represents the empirical average reward since the logging and target policy are the same. 
We expect the off-policy estimators to recover the true average reward of the policy as indicated by \emph{online} scores in each of the plots. 
Each experiment was repeated 20 times and the error bars represent 95\% confidence intervals.

We observed that the~\emph{Standard IPS} estimator has a large variance, making it less favorable to be used in practice.~\emph{ IIPS}, on the other hand, has the least variance but provides a biased estimate whereas RIPS has considerably lower variance than~\emph{Standard IPS} and provides the most accurate estimate amongst the three. %

\subsection{How do \emph{ESS} threshold and slate size affect the proposed RIPS estimator?}
\begin{figure}[t!]
\centering
\includegraphics[scale=.2]{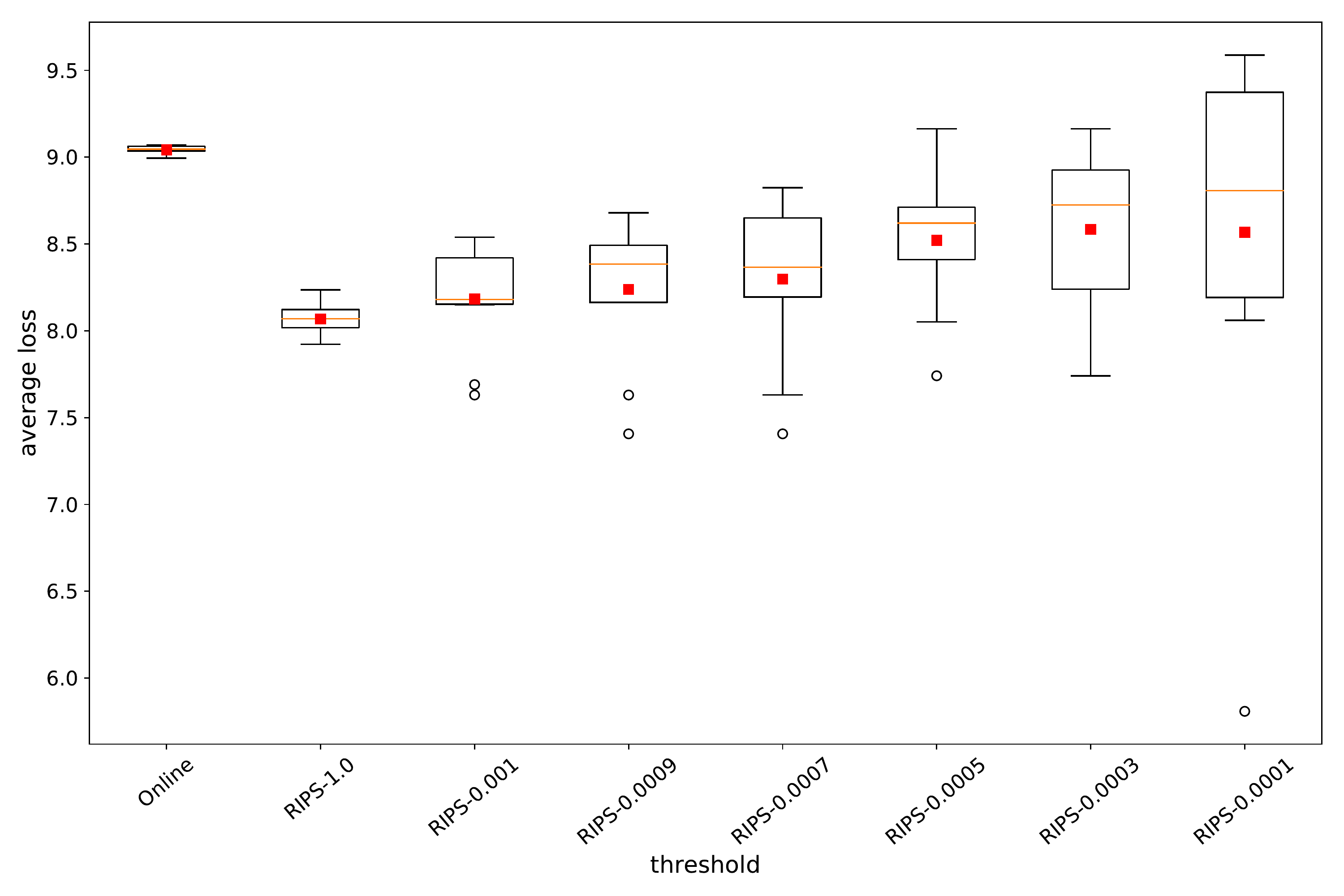}
\caption{Performance of the RIPS estimator for different threshold values (closer to \texttt{Online} is better).}
\label{fig:threshold_plot}
\end{figure}

The threshold parameter $t$ used in the RIPS estimator controls the lookback size for a given slate to mitigate extreme weighting.
The threshold is a way of reducing the variance of the estimator while introducing a small amount of bias. 
To better understand the effect of the threshold parameter on the performance of RIPS estimator, 
we set up an experiment to evaluate RIPS under different threshold values.
Figure~\ref{fig:threshold_plot} shows the average loss of the \emph{anti-optimal} policy estimated by the RIPS estimator with different threshold values and compares it to the empirical average as indicated by \emph{online}.  

When threshold $t=1.0$, RIPS decomposes to~\emph{IIPS} since there is no lookback, i.e., \emph{line 12} in Algorithm~\ref{alg:RIPS} will never be true.
Next, we increasingly vary the threshold~\footnote{We ignore values $t=1.0$ to $t=0.001$ because at those thresholds effective sample size was zero resulting in the same average reward value as RIPS-$1.0$. } to demonstrate the effect on the mean and variance of the estimates. 
It can be observed that as we relax the threshold the RIPS estimator looks back further to consider the reward interaction. 
The reason for the increased variance is due to fewer data available which results in extreme weighting. 
Since the threshold parameter directly depends on the number of effective samples, we wish to explore methods to enable automatic tuning of the threshold parameter in the future. 

\begin{figure*}[t!]
\centering
\includegraphics[scale=.40]{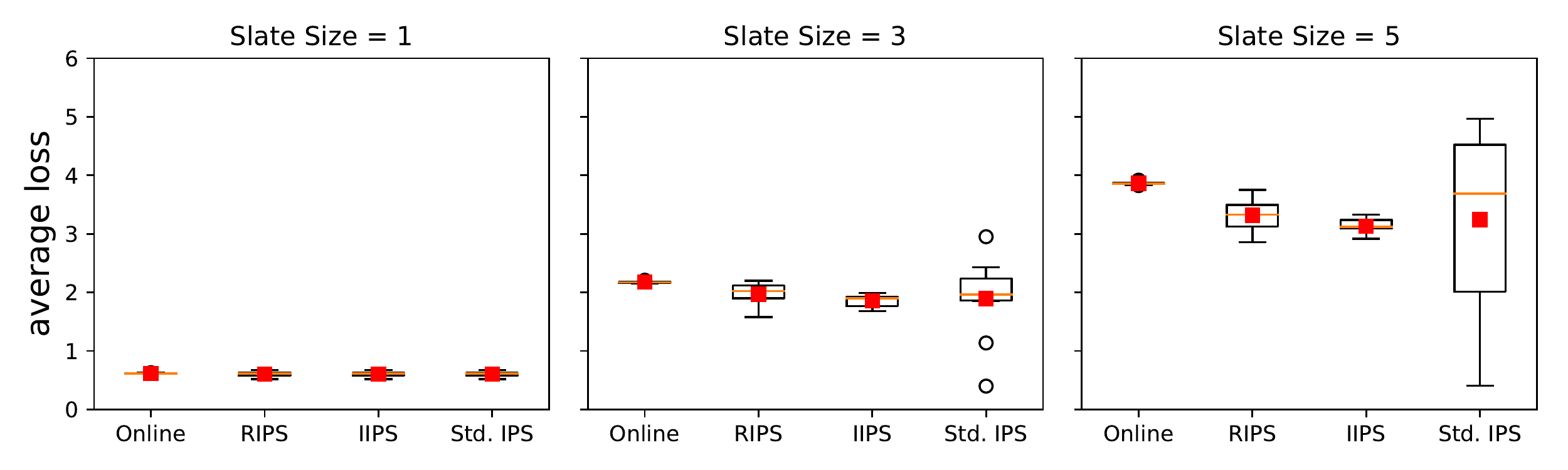}
\caption{Performance of the estimators for different slate size.}
\label{fig:slate-size}
\end{figure*}

Finally, we repeat the simulation experiment for different slate sizes with the random policy as the logging and optimal policy as the target policy. Figure~\ref{fig:slate-size} shows the average reward according to different estimators. When the slate size is one, all estimators behave similarly as there are no reward interactions in this case but as the slate size increases, we notice a gradual increase in the variance of the~\emph{Standard IPS estimator} and the bias in~\emph{IIPS} increases. RIPS is least affected by the slate size.  
    
\section{Experiments on Real-World Data}
\label{sec:realexperiments}

In this section, we investigate how well various off-policy estimators perform at evaluating track sequencing algorithms used in an online environment to shuffle tracks within playlists on a large music streaming platform. 
An A/B test was run over a 6-week period in the summer of 2019 using 4 different shuffling algorithms: (1) a uniformly random shuffle, (2) a shuffle algorithm biased towards selecting more popular tracks first and which is non-personalized, (3) a personalized shuffle algorithm biased towards selecting tracks that the user has the highest affinity towards as predicted by a proprietary user-track affinity model, and (4) a shuffle algorithm biased towards selecting tracks acoustically similar to the previous track the user has heard. 
Note that policies (1), (2) \& (3)  do not depend on previous actions but (4) does.
Approximately four million users, chosen at random, were exposed to one of these variants at random when commencing a session listening to one of a subset of 15 popular playlists available to all users on the platform.
We define a session to be a continuous period of listening on a single playlist with no pause of greater than 60 seconds between playback of subsequent tracks.

In our study, we consider two primary metrics to compare different sequencing algorithms: skip rate and the listening time per session.
Skip rate is the ratio of number of track skipped in a session to the number of tracks played and listening time is the total time spent in a session. 
Table~\ref{tab:onlineabtest} summarizes the online performances of the four policies tested in this study as characterized by the inverse of skip rate and listening time per session. 
As shown in the table, the affinity-biased approach has both the fewest skipped tracks in the first 10 of the session and the longest session length, followed closely by popularity-biasing.  Acoustic biasing produces fewer skips than uniform random shuffling, but also slightly shorter listening sessions.

\subsection{Comparison of Off-Policy Estimators}
We measure how well each estimator predicts the online outcomes, based on logged data collected under each of the four shuffling algorithms, i.e., \emph{logging policies}. 
We use skip-rate as the primary metric to be predicted, cap the session length to a maximum of 10 tracks, and compute our estimates using data collected under the uniformly random logging policy. We evaluate our estimators in terms of RMSE with respect to the ground truth obtained on-policy during the A/B test, that is, RMSE computed over the four estimates produced by each estimator.  We compare our estimator, RIPS, against baselines described in~\ref{sec:exp:offpolicy}

\begin{table*}[t]
\begin{center}
  \begin{tabular}{ | c | c | c | c | }
    \hline
    Sequencer & Unskipped tracks in top 10 & Session length & Number of users \\ \hline
    Uniform random& $5.86$ & $701.7$s & $1,012,658$  \\ \hline 
    Popularity biased & $6.00$ & $726.9$s & $1,016,195$ \\ \hline 
    User affinity biased& $6.09$ & $733.3$s & $1,010,027$  \\ \hline 
    Acoustic biased & $5.90$ & $697.1$s & $998,708$ \\ \hline 
    \hline
  \end{tabular}
\end{center}
\caption{Results and number of users for each sequencer in the online A/B test. Our primary metric is the number of unskipped tracks in the first 10 tracks of a session. All means for results significantly different at $0.05$ level according to Tukey's HSD test for comparing pairs of means with corrections for multiple comparisons.}
\label{tab:onlineabtest}
\end{table*}

Each of the off-policy estimators is compared using data collected under the uniformly random logging policy. 
Since deploying a uniform random logging policy for an extended period is highly undesirable due to the negative impact it has on business metrics, it is important to compare the effectiveness of the estimators under non-uniform logging policies too.
Table~\ref{tab:loggingpolicy} shows the RMSE for each estimator, we observe that RIPS outperforms the baselines and the advantage for RIPS is even greater under the non-uniform logging policies. 
The fact that IIPS shows a stronger decline in performance for the other logging policies may reflect that the sequential dependencies between rewards are stronger when the logging policy is not simply uniformly sampling from the available track pool but is instead biased towards a particular subset of the pool. 

While Table~\ref{tab:loggingpolicy} used the skip-rate metric to compare policies, we additionally compared the estimators using the listening time per session (in seconds) metric. The results followed a similar trend, we observed that the performance of RIPS was even more pronounced with an RMSE of \textbf{43.3} compared to IIPS (63.2) and IPS (303.0). 

\begin{table}[t]
\begin{center}
  \begin{tabular}{ | c | c | c | c | c | }
    \hline
    Logging policy  & IPS & IIPS & RIPS ($\%\Delta$ from IIPS) \\ \hline
    Uniformly random & 1.893 & 0.263 & \textbf{0.194} (26\%)\\ \hline 
    Popularity biased & 1.812 & 0.681 & \textbf{0.391} (43\%) \\ \hline 
    User affinity biased & 1.932 & 0.284 & \textbf{0.196} (30\%) \\ \hline 
    Acoustic biased & 3.480 & 0.370 & \textbf{0.364} (2\%) \\ \hline 
    \hline
  \end{tabular}
\end{center}
\caption{RMSE of estimators under the different logging policies.}
\label{tab:loggingpolicy}
\end{table}

\subsection{Impact of Dataset Size}
\label{sec:exp:size}

\begin{figure}[t!]
\centering
\includegraphics[width=0.85\linewidth]{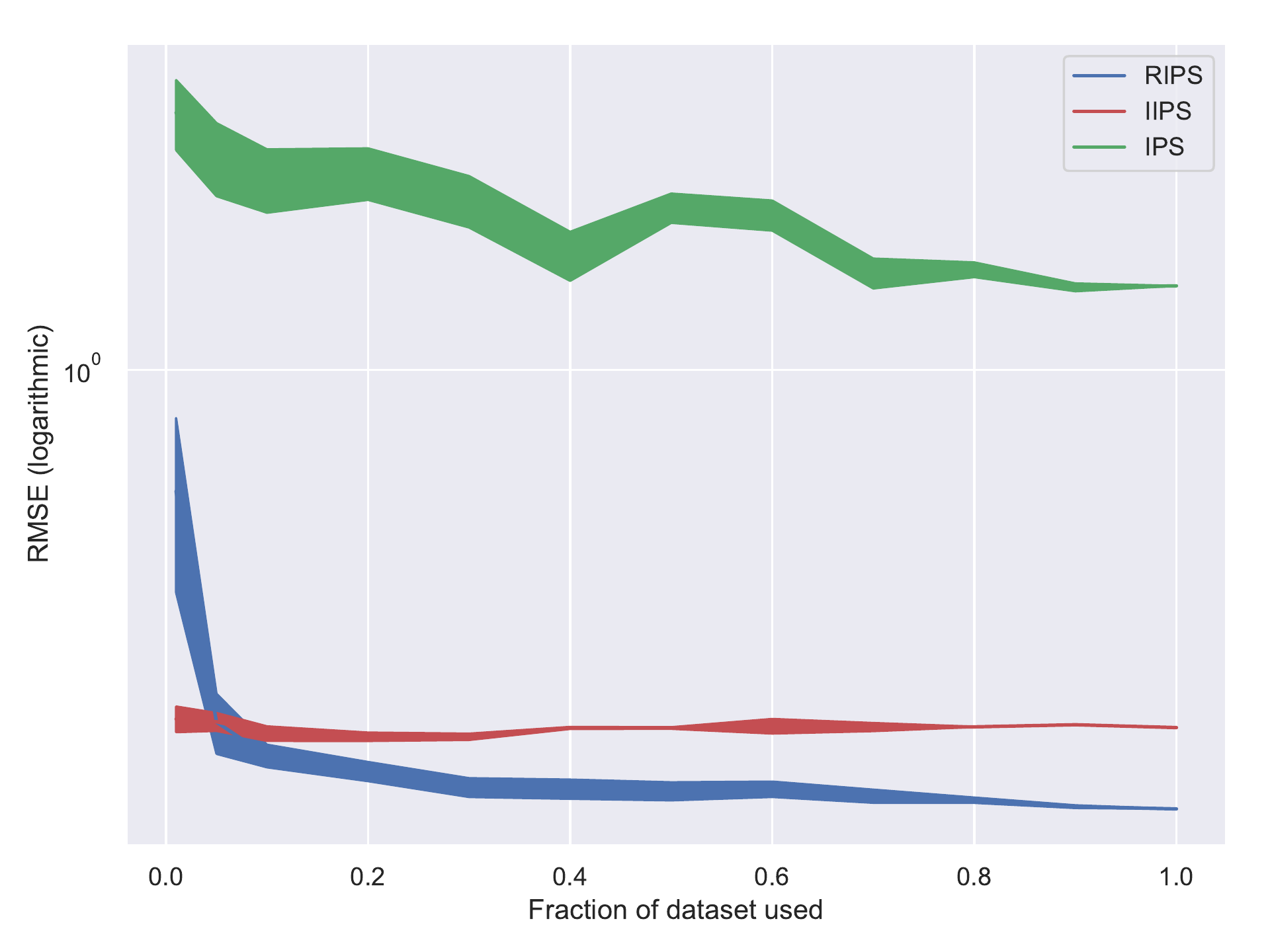}
\caption{Performance of the estimators for different dataset sizes. Note that the $y$ axis is logarithmic to allow the results for IPS to be shown clearly alongside the differences between IIPS and RIPS. The shaded areas indicate the maximum and minimum across 5 runs for each estimator}
\label{fig:data-size}
\end{figure}

We begin by investigating how the estimators perform for different dataset sizes. We vary the proportion of our dataset available to the estimators, from 1\% of the data, corresponding to approximately 50,000 sessions, to 100\% of the data, i.e. all 5 million sessions collected under the uniformly random logging policy. The results, shown in Figure~\ref{fig:data-size}, demonstrates the poor performance of IPS and that once we use more than 15\% of the logged data, RIPS significantly outperforms IIPS. IIPS does not improve noticeably as the size of the dataset increases, suggesting that bias, rather than variance is the issue. Conversely, for IPS and RIPS, increasing the amount of data leads to better estimates. For the full dataset, RIPS obtains an RMSE of 0.194, compared to an RMSE of 0.263 for IIPS, and 1.893 for IPS. Note that the $y$ axis is logarithmic to allow the results for IPS to be shown alongside those of IIPS and RIPS.

\subsection{Impact of Slate Size}
\label{sec:exp:length}

\begin{figure}[t!]
\centering
\includegraphics[width=0.85\linewidth]{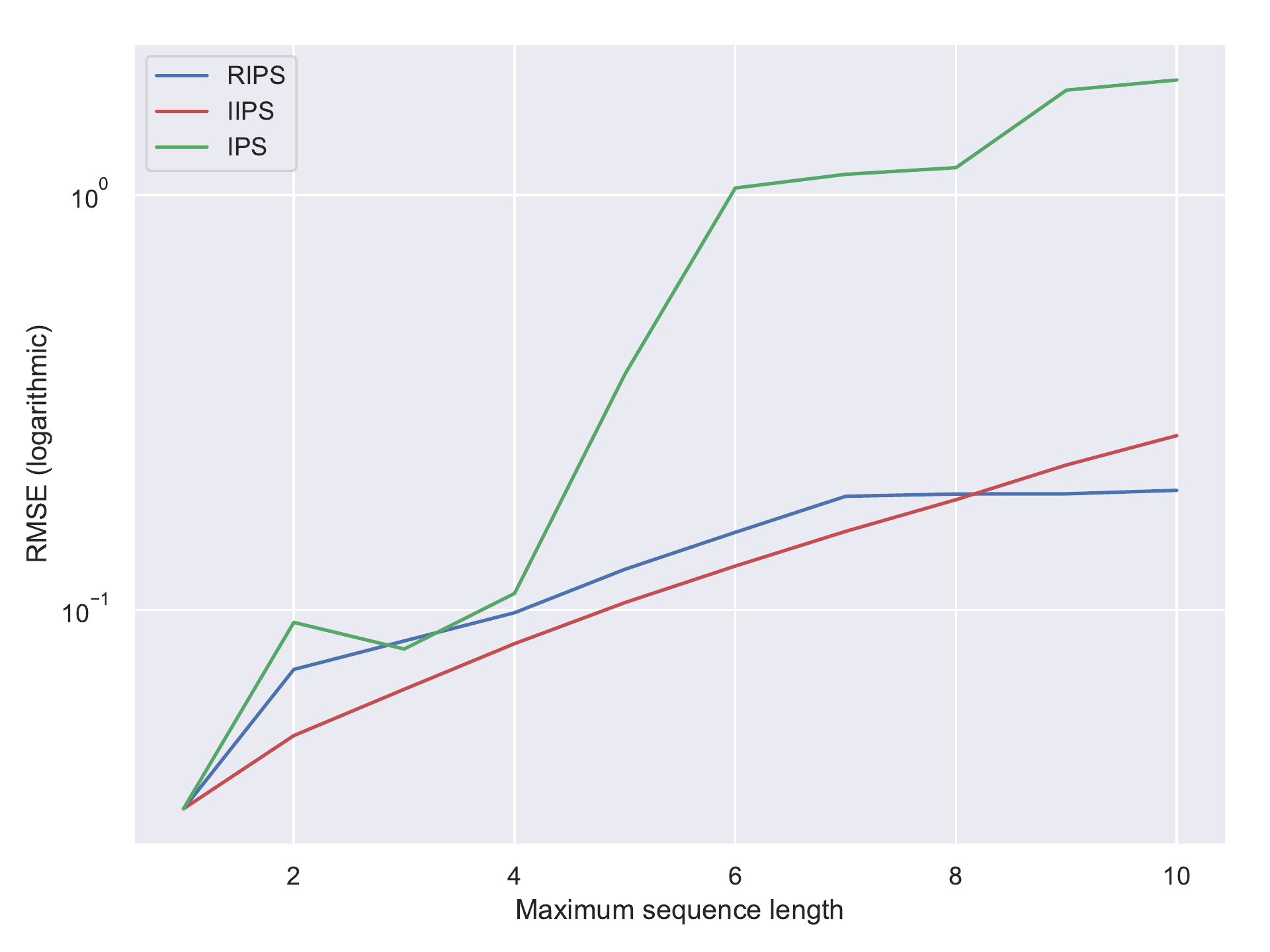}
\caption{Performance of the estimators for different slate sizes (i.e., sequence lengths). Note again the log scale on the $y$ axis.}
\label{fig:sequence-length}
\end{figure}

Since the propensities grow exponentially as the lengths of the slate size increase, an estimator like IPS is expected to perform poorly for longer sequences. 
When does this effect become prohibitive, and when do potential sequential dependencies have enough impact for IIPS to become biased? 
Figure~\ref{fig:sequence-length} shows how the estimators perform when we consider different slate sizes (i.e., when we consider only those sessions with shorter than a given threshold number of tracks). 
 We note that even for slate size as short as length 5, the RMSE for IPS begins to diverge significantly. For IIPS, the bias caused by the independence assumption does not result in substantially worse performance than for RIPS until we get to size 8 or longer.

\subsection{Comparison with Pseudoinverse Estimator }
\label{sec:exp:PI}

For completeness, even though the assumptions of the pseudoinverse~(PI) estimator are violated in our problem setting, as noted in Section~\ref{ss:pi}, we include an experimental comparison with PI~\cite{swaminathan2017off}.
We use the simplified version described in Example 5 in~\cite{swaminathan2017off}, i.e., the case when the logging policy is uniform random, and the slate size is equal to the number of candidates. Here, we observe an RMSE of 0.738 for IIPS, 0.740 for PI, and \textbf{0.291} for RIPS. We omit the generalized form of PI proposed in~\cite{swaminathan2017off} (Eq. 5 and 6), since there is no closed-form solution for computing $\Gamma$ for our logging policies.

\section{Conclusions and Future Work}
\label{sec:Conclusion}

In this paper, we presented RIPS, a new counterfactual estimator for sequential recommendations derived from a causal graph specification of the interactions between actions and rewards in a slate. 
Our approach is a lightweight non-parametric method that avoids the overhead of full parametric estimation associated with the direct method and is asymptotically unbiased. 

Experiments using simulations and on a live system validate that RIPS has lower variance than IPS and lower bias than independent IPS when sequential reward interactions exists. 
This translates to a lower error when estimating slate-level skip rates and listening time for playlists.  
This improvement is maintained under different non-uniform logging policies, a key application for offline evaluation. 
Finally, our findings also indicate that significant sequential reward interaction exists among songs within playlists.

In future work, the methodology presented here may be extended to other kinds of reward interactions, e.g., a 2-dimensional grid of actions or longer-range dependencies. 
Another promising area to explore in the future merging our model agnostic approach with recent advances in doubly robust estimation such as \cite{thomas2016data, su2019cab, bibaut2019more}

\begin{acks}
We are very grateful to the anonymous reviewers for their constructive feedback, Jose Falcon for his help in creating the experimental infrastructure, and Thor Kell and Frej Connolly for their help in setting up the A/B Test. 
\end{acks}

{\small
\bibliographystyle{abbrv}
\bibliography{Bib} }

\appendix

\section{Properties of RIPS}
\label{appendix:rips}

\begin{algorithm}[t!]
	\label{alg:RIPS}
 \SetKwInOut{Input}{input}
 \SetKwInOut{Output}{output}   
 \Input{ target policy $h$, logging policy $\pi$, threshold factor $t$, observed contexts $X^{(1:N)}$, actions $A^{(1:N)}_{1:K}$, and rewards $R^{(1:N)}_{1:K}$}
 \For{$\mathrm{position\;\;} k \gets 1$ \KwTo $K$}
 {
	 \For{$\mathrm{sample\;\;} n \gets 1$ \KwTo $N$}	
	 {
		calculate independent weight $w_k^{(n)} \gets \frac{h(A_k^{(n)} \g X^{(n)})}{\pi(A_k^{(n)} \g X^{(n)})}$ 
	}
	initialize lookback $b_k \gets 0$\\
	initialize RIPS weights $\gamma_k^{(1:N)} \gets 1$\\
	\While{$b_k < k$}
	{
		\For{$\mathrm{sample\;\;} n \gets 1$ \KwTo $N$}	{
			calculate proposal weight $u_k^{(n)} \gets \frac{N \gamma_k^{(n)} w_{k - b_k}^{(n)}}{\sum_{n'=1}^N \gamma_k^{(n')} w_{k- b_k}^{(n')}}$
		}
		
		using Eq.~\ref{eq:ESS}:\\
		
		\lIf{$\mathrm{ESS}(u_k^{(1:N)}) > N t \mathrm{\;\;and\;\;} \mathrm{ESS}(u_k^{(1:N)}) < \mathrm{ESS}(\gamma_k^{(1:N)})$}
			{ $\gamma_k^{(1:N)} \gets u_k^{(1:N)}$  \DontPrintSemicolon}
		\lElse
			{ break \DontPrintSemicolon}
		$b_k \gets b_k+ 1$
	} 
 }
 \Return target policy value estimate $\hat{V}(h) \gets \frac{1}{N} \sum_{n=1}^N \sum_{k=1}^K \gamma_k^{(n)} R_k^{(n)}$
 \caption{Reward interaction inverse propensity scoring (RIPS)}
 \label{alg:RIPS}
\end{algorithm}

The appendix provides a discussion of the properties of reward interaction inverse propensity scoring (RIPS). 
The pseudocode for RIPS is provided in Algorithm~\ref{alg:RIPS}. 
First, the bias tends to zero as $N \rightarrow \infty$ 
assuming that the rewards have the conditional dependencies 
in Fig.~1c. 
Second, RIPS has lower variance 
than standard IPS with slate data in Eq.~\ref{eq:ips}.

\emph{RIPS Unbiased in the Limit of Infinite Data:} Need to prove that $\E[ \gamma_{k-1}^{(n)} w_k^{(n)} ] = 1$. Proof by induction:

\begin{itemize} 
	\item \emph{Base case}, $k=1$:  $\E[ 1 \times w_k^{(n)} ] = 1$ by the derivation in Eq.~\ref{eq:derive_1}. %
	\item \emph{Inductive case}, $k \rightarrow k+1$: recall that $\gamma_{k-1}^{(n)}$ is a random variable depending on $A_{1:k-1}$ and $w_k^{(n)}$ is a random variable depending on $A_k$, so make these dependencies explicit (for slate~$n$),
	\begin{flalign}
		&\E[ \gamma_{k-1}^{(n)}(A_{1:k-1}) w_k^{(n)}(A_k) ] \\
		&= \int \gamma_{k-1}^{(n)}(A_{1:k-1}) w_k^{(n)}(A_k) \prod_{j=1}^k \pi(A_j \g A_{j-1}, X) \mathrm{d}A_{1:k} %
	\end{flalign}
	\begin{flalign}
		\;&= \int \gamma_{k-1}^{(n)}(A_{1:k-1}) h(A_k \g A_{k-1}, X) \prod_{j=1}^{k-1} \pi(A_j \g A_{j-1}, X) \mathrm{d}A_{1:k} \\
		&= \int \gamma_{k-1}^{(n)}(A_{1:k-1}) \prod_{j=1}^{k-1} \pi(A_j \g A_{j-1}, X)  \left( \int h(A_k \g A_{k-1}, X) \mathrm{d}A_k \right) \mathrm{d}A_{1:k-1}\\
		&= \int \gamma_{k-1}^{(n)}(A_{1:k-1}) \prod_{j=1}^{k-1} \pi(A_j \g A_{j-1}, X) \mathrm{d}A_{1:k-1}\\
		&= 1.
	\end{flalign}
\end{itemize}
where inductive hypothesis was used in the final line. $\blacksquare$

By the strong law of large numbers, 
the sampled weights will be equal to the population mean 
as $n \rightarrow \infty$. In the limit, the weighting factor in Eq.~7 is equal to $\frac{1}{N}$ and $\hat{V}_\mathrm{RIPS}$ reduces to 
IPS which is unbiased.

\paragraph{Variance Properties of RIPS}

We compare the variance of RIPS 
with that of the NIS applied to slate estimation \cite{gilotte2018offline}. 
NIS has favourable variance reduction properties 
and is the most suitable comparison 
due to its widespread use. 
\begin{flalign}
	\hat{V}_\mathrm{NIS}(h) &:= \frac{1}{N} \sum_{n=1}^N \frac{\prod_{k=1}^K w_k\n}{\sum_{m=1}^N \prod_{k=1}^K w_k^{(m)}} \sum_{k=1}^K R_k\n \label{eq:nis}
\end{flalign}

To aid the comparison, 
here is the closed-form RIPS expression,
\begin{flalign}
	\hat{V}_\mathrm{RIPS} &= \sum_{n=1}^N \sum_{k=1}^K \frac{\prod_{k'=1}^k w_{k'}\n} { \sum_{m=1}^N \prod_{k'=1}^k w_{k'}^{(m)}} R_{k}\n, \label{eq:closed_rips}
\end{flalign}
which can be seen by the fact that 
the denominator in Eq.~7 for position $k$ 
cancels out the denominator 
for position $k-1$ leaving only the sum-product 
of all the weights up to 
and including position $k$ as the normalizer.  

Here is a proof by induction that 
RIPS has non-strictly lower variance than NIS. 
Comparing Eq.~\ref{eq:ips} with Eq.~\ref{eq:closed_rips}, 
notice that both sum over the 
reweighted position-based rewards $R_k\n$ 
for $k=1,\dots,K$. 
We start at the last position and work backwards. 
\begin{itemize}
	\item \emph{Base case}: starting at position $K$ in the slate, 
	the reweighting factor for both RIPS and NIS 
	is $\prod_{k=1}^K w_k\n$, therefore the variance of the 
	reward estimator for the last position is identical. 
	\item \emph{Inductive case}: moving from position $k \rightarrow k-1$, 
	observe that the number of reweighting factors in RIPS reduces 
	by 1 while the number of reweighting factors in NIS remains constant. 
	By the fact that $\mathrm{Var}\left[\frac{h(A_k \g X)}{\pi(A_k \g X)} \right] \ge 0$, 
	the variance of RIPS must be strictly less than that of NIS. 
\end{itemize}
Further, RIPS has a strictly lower variance than NIS 
if any of the reweighting factors $k < K$ 
has non-zero variance, 
which is commonly the case. 
$\blacksquare$

Finally, as $N \rightarrow \infty$, the effective sample size also goes to infinity if absolute continuity between the logging and target policies holds.

\end{document}